\documentclass[a4paper,10pt,conference]{ieeeconf}
\IEEEoverridecommandlockouts
\usepackage{graphicx}
\usepackage{times}
\usepackage{amsmath}
\usepackage{amssymb}
\usepackage{color}
\usepackage[export]{adjustbox}
\usepackage{booktabs}
\usepackage{multirow}
\usepackage{subfig}
\usepackage{epstopdf}
\usepackage{avldefs}
\usepackage[margin=20mm]{geometry}

\usepackage{comment}
\usepackage[table]{xcolor}
\def\ie{\emph{i.e.} }
\def\eg{\emph{e.g.} }

\newcommand\hl[1]{\textbf{#1}}

\title{Cos R-CNN for Online Few-shot Object Detection}

\author{\authorblockN{G. Wesley P. Data, Henry Howard-Jenkins, David W. Murray, Victor A. Prisacariu}
\authorblockA{Active Vision Laboratory, Department of Engineering Science \\
University of Oxford \\
Oxford OX1 3PJ, United Kingdom}}

\begin{document}

\maketitle

\begin{abstract}

We propose Cos R-CNN, a simple exemplar-based R-CNN formulation that is designed for online few-shot object detection. That is, it is able to localise and classify novel object categories in images with few examples without fine-tuning. Cos R-CNN frames detection as a learning-to-compare task: unseen classes are represented as exemplar images, and objects are detected based on their similarity to these exemplars. The cosine-based classification head allows for dynamic adaptation of classification parameters to the exemplar embedding, and encourages the clustering of similar classes in embedding space without the need for manual tuning of distance-metric hyperparameters. This simple formulation achieves best results on the recently proposed 5-way ImageNet few-shot detection benchmark, beating the online 1/5/10-shot scenarios by more than 8/3/1\%, as well as performing up to 20\% better in online 20-way few-shot VOC across all shots on novel classes.

\end{abstract}

\section{Introduction}

Deep data-driven feature learning has enabled rapid advances in visual recognition tasks such as image classification~\cite{NIPS2012_4824,Simonyan2015,7298594,He_2016_CVPR} and object detection~\cite{NIPS2015_5638,8237586,He_2017_ICCV,redmon2018yolov3} due to its ability to learn good features from data. However, training deep models from scratch requires a large amount of training data, which makes its application in a potentially much larger number of data-scarce tasks prohibitive. To this end, the problem of learning from few labelled annotations, called few-shot learning, needs to be tackled. This is reflected in an ever-growing body of work studying few-shot image classification~\cite{Lake1332,NIPS2016_6385,NIPS2016_6068,NIPS2017_6996,pmlr-v70-finn17a,Ravi2017,DBLP:journals/corr/abs-1803-02999,rusu2018metalearning,Sung_2018_CVPR}.

Despite rapid progress in few-shot image classification, the analogous task of few-shot object detection has received comparatively less attention~\cite{AAAI1816778,Karlinsky_2019_CVPR,Kang_2019_ICCV,Yan_2019_ICCV,Wang_2019_ICCV,wang2020frustratingly,Perez-Rua_2020_CVPR}. Furthermore, most of existing literature~\cite{AAAI1816778,Kang_2019_ICCV,Yan_2019_ICCV,Wang_2019_ICCV,wang2020frustratingly} rely on a 2-stage training pipeline that involves 1) training a base detector using abundant examples of base classes, and then 2) fine-tuning the base detector on data-scarce novel class examples. While this approach has been sucessful, it is not appropriate for deployment in resource-constrained applications such as embedded devices as these devices typically have limited computational capacity, which may prohibit the training of large networks. Furthermore, these settings require base class examples to be available during fine-tuning to prevent catastrophic forgetting~\cite{french1999catastrophic}, which again may prohibit fine-tuning if the devices are memory-constrained.

\begin{figure*}
  \centering
  \includegraphics[width=0.99\textwidth]{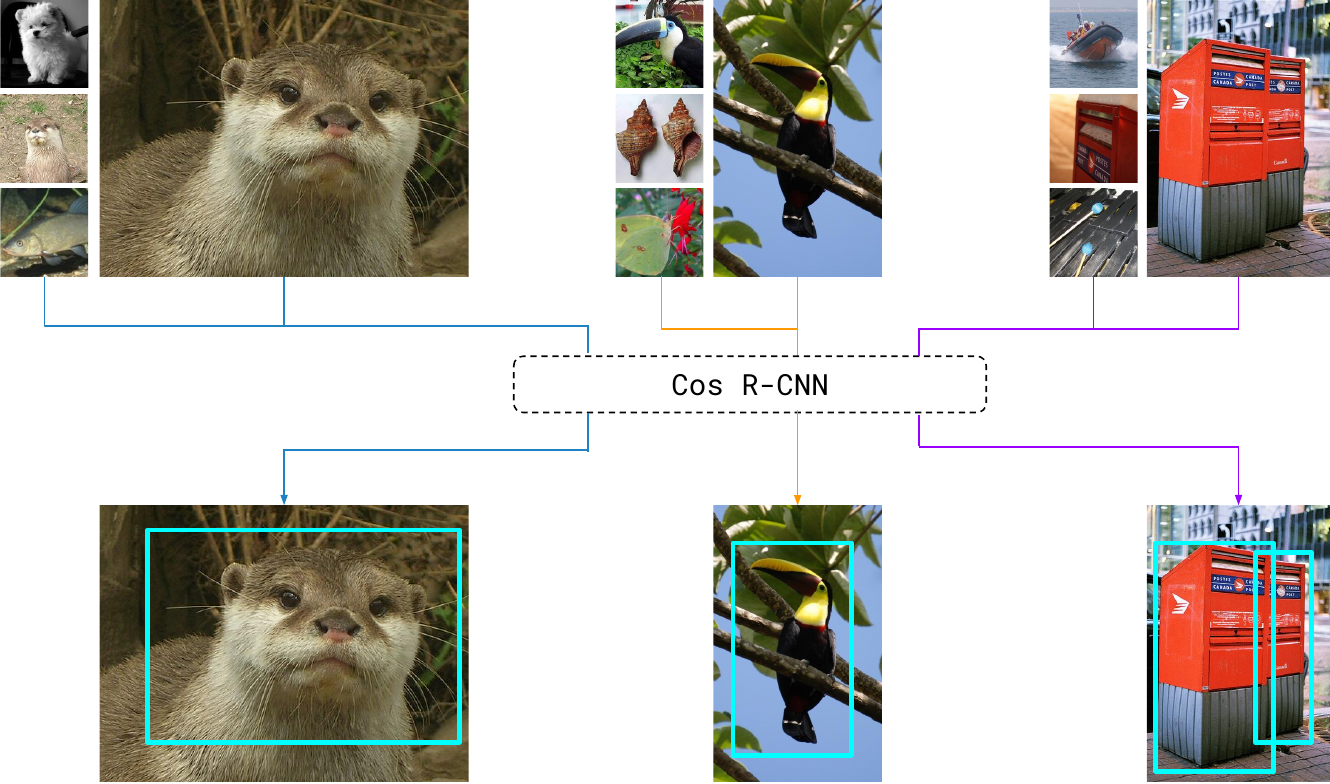}
  \caption{Three sets of distinct 3-way, 1-shot detection tasks. For each task, a query image containing a novel object is to be detected by providing exemplar images, at least one of which contains the novel object. Cos R-CNN is a online few-shot object detector, meaning that the same model can perform all 3 tasks having to be fine-tuned.}
  \label{fig:intro}
\end{figure*}

It is then useful to study few-shot detection in an online setting, \ie to be able to train a model that can detect unseen classes without having to be fine-tuned as illustrated in Fig.~\ref{fig:intro}.

Our approach, Cos R-CNN, follows the general structure of an R-CNN, but is trained to generate locations for objects that are similar to exemplars of known labels. This naturally avoids the need for fine-tuning, as the same model can be used for different sets of detection classes. Inspired by recent approaches from few-show classification~\cite{chen2018a,Gidaris_2018_CVPR,Qiao_2018_CVPR} we use cosine distance as our similarity measure, as it scales the logits (\ie pre-softmax) of any class into the same range. Additionally, we also intuit that cosine-based metric learning may benefit from being shown a large quantity of data in each iteration (i.e. having large batch sizes), but this is usually difficult to obtain as R-CNN based frameworks possess a large memory footprint during training even at small batch sizes (\eg 1-2). To work around this, we also introduce Su-MoCo, a variant of MoCo~\cite{he2019momentum} that enables a much larger corpus of exemplars to be compared to, improving the learning process.

We structure our paper as follows. In \S\ref{sec:related} we summarise related work on few-shot object detection and classification. In \S\ref{sec:method}, we detail each component of Cos R-CNN, and in \S\ref{sec:experiments} then evaluate our model, benchmarking it on the 5-way detection task introduced in~\cite{Karlinsky_2019_CVPR}, and detail ablations for our design choices. Finally, we summarise our findings in \S\ref{sec:conclusion}.

\section{Related Work}
\label{sec:related}

We explore the background of few-shot research that provides context for this paper. Object detection combines the traditional task of image classification with an additional constraint of localisation. In other words, instead of classifying an image as a whole, the class and extent of an arbitrary number of object regions must be predicted. While there have been a number of successful detectors~\cite{NIPS2015_5638,10.1007/978-3-319-46448-0_2,8100173,redmon2018yolov3,8237586}, few-shot object detection remains relatively under-explored. Therefore, we initially explore work towards few-shot image classification, which is closely related to object detection. Later, we detail recent methods for few-shot object detection.

\subsection{Few-shot Classification}
\label{sec:fewshot_review}
Few-shot learning is a form of supervised learning that, in contrast to deep learning, aims to achieve generalisation using few examples. There has been increasing interest recently in few-shot learning. \cite{Lake1332} proposed a Bayesian framework for recognising unseen handwritten characters by distilling them into strokes and combining them in an example-efficient manner. In the same year, \cite{Koch2015} used a deep Siamese convolutional network to determine if pairs of known/unknown image exemplars belong to the same class, framing few-shot classification as a matching task.

A concept that arose relatively recently is meta-learning, where models are set up to learn over a distribution of few-shot tasks by making it iterate over ``episodes" of few-shot tasks. This is unlike standard supervised learning where training is performed on batches of examples which make up a single task. By learning distributions of few-shot tasks, the model aims to generalise to any possible future few-shot task. This is first demonstrated by Matching Networks~\cite{NIPS2016_6385}, which trained an exemplar matching network based on k-NN cosine similarity using this paradigm.

Similarity or matching-based strategies have since then been featured in a growing number of works. Prototypical Networks~\cite{NIPS2017_6996} compared Euclidean distances of query embeddings to meta-learned class prototypes and matched the closest pair. Relation Networks~\cite{Sung_2018_CVPR} expanded on this idea and made the similarity measure learnable by introducing a relation module to do the comparison.

Other works such as MAML~\cite{pmlr-v70-finn17a}, Meta-SGD~\cite{DBLP:journals/corr/LiZCL17}, and Reptile~\cite{DBLP:journals/corr/abs-1803-02999} expand the meta-learning strategy by inventing solvers to learn a network initialisation that is amenable to the few-shot context. In the case of meta-learner LSTM~\cite{Ravi2017}, an SGD update rule is additionally also learned to guide few-shot optimisation.

More recently, works such as~\cite{Gidaris_2018_CVPR,Qiao_2018_CVPR,chen2018a} demonstrated that, given a powerful enough feature extractor, a scaled cosine similarity formulation for classification is competitive with many state-of-the-art few-shot methods.

\subsection{Few-shot Object Detection}
\label{sec:fewshot_detection_review}
The literature in few-shot object detection is sparser when compared to few-shot classification. Most methods propose a 2-stage training pipeline where a base detector is first trained on abundant base classes, and then fine-tuned on scarce novel classes. Amongst the first to adopt this is LSTD~\cite{AAAI1816778}, where they proposed to regularise the fine-tuning stage with additional objectives designed to reduce the severity of overfitting to the scarce novel classes. More recently, FSDet~\cite{wang2020frustratingly} showed that a simple strategy of freezing weights and fine-tuning only the last box head layer of an R-CNN detector is competitive with many contemporary few-shot object detection methods.

Still following the general 2-stage pipeline, other works propose to generate network weights for classifying novel classes in the second stage. Fewshot-YOLOv2~\cite{Kang_2019_ICCV} and Meta-RCNN~\cite{Yan_2019_ICCV} aim to modulate base network features with channel-wise attention vectors generated by a subnetwork from the few-shot exemplars. Meta R-CNN additionally introduced a loss function that encourages these vectors to semantically cluster in the attention vector space. In a similar fashion, MetaDet~\cite{Wang_2019_ICCV} also learns a subnetwork to predict parameters, but directly treats them instead as classification weights instead of using it to modulate incoming features.

In contrast to previous methods, RepMet~\cite{Karlinsky_2019_CVPR} proposed the usage of distance metric learning to learn an embedding space for object categories, each of which is represented by a multi-modal distribution. This could then be applied to few-shot detection by replacing the network head in an R-CNN with the learned object embeddings. Classification is performed using the distance of the query from representative embeddings of novel categories to explicitly compute class posteriors.

Similar to RepMet, the cosine formulation of Cos R-CNN can be thought of to induce metric learning, which can then be exploited to perform online detection. Unlike RepMet, however, Cos R-CNN is based on pure pairwise comparison, simplyfing the formulation while also surprisingly (as we can see in \S\ref{sec:experiments}) obtaining better performance. Furthermore, Cos R-CNN is also much simpler with less training objectives and having no specific hyperparameters to tune.

\section{Cos R-CNN}
\label{sec:method}

\begin{figure*}
  \centering
  \includegraphics[width=\textwidth]{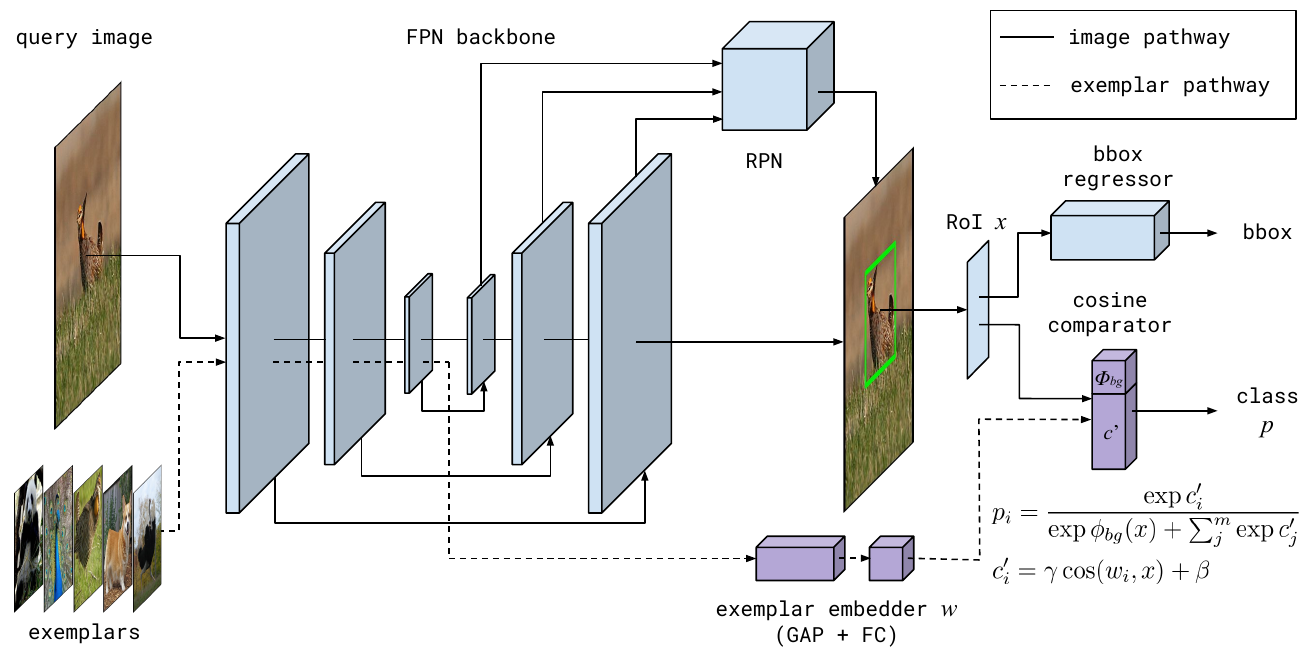}
  \caption{Overview of Cos R-CNN. Components which are unaltered from the original Faster R-CNN are coloured in blue, while those modified to enable few-shot detection are coloured in purple.}
  \label{fig:method}
\end{figure*}

This paper's proposed architecture is shown in Figure~\ref{fig:method}. It roughly follows the form of a generic, two-stage, R-CNN approach. However, in a standard R-CNN detection approach, the model learns to detect object categories over many iterations and, importantly for generalisation, with many different examples of the same object type. This formulation is ill-suited for few-shot detection: any attempt to train for novel class detection without careful regularisation will likely overfit to the small number of specific examples. This overfitting is only exacerbated in our few-shot scenario where there are only few training examples.

Cos R-CNN tackles this issue of single-example generalisation within an object category with cosine similarity, which explicitly encourages the network to learn an embedding that clusters similar categories together. Intuitively, this formulation achieves a similar objective to a metric learning formulation such as in~\cite{Karlinsky_2019_CVPR} without having to tune a margin hyperparameter.

Specifically, the structure of Cos R-CNN roughly follows that introduced by Faster R-CNN~\cite{NIPS2015_5638}. The first stage is a sub-network called the Region Proposal Network (RPN), which proposes candidate objects and their bounding boxes. We refer to the second stage as the network \emph{head}. It is a small sub-network which aims to classify candidate objects and to refine their bounding boxes. The network head operates on fixed-sized feature maps that are pooled from an image feature map through RoIAlign~\cite{He_2017_ICCV}. The head is therefore essentially Fast R-CNN~\cite{Girshick_2015_ICCV}. Inference time is reduced by having the RPN and head operate on the same image feature map, computed by a convolutional \emph{backbone}. Commonly, ResNets~\cite{He_2016_CVPR} with a Feature Pyramid Network (FPN)~\cite{Lin_2017_CVPR} are used as the backbone architecture.

\subsection{Exemplar Embedding Pathway}
\label{sec:exemplar_method}
As the cosine similarity formulation necessitates comparisons of the query image with exemplars, the typical R-CNN pipeline needs to be modified to generate these exemplars. We choose to generate exemplar features by piping images of cropped exemplar instances through the ResNet-FPN backbone, followed by an exemplar embedder as shown in Figure~\ref{fig:method}. The exemplar embedder consists of a global average pooling (GAP) layer followed by a fully-connected layer that maps the pooled feature vector of channels 2048~$\rightarrow$~1024 so that it matches the channel dimensions of RoI features from the query image, making them comparable by cosine similarity. We also note that sharing the backbone in this manner allows exemplar features to be generated without substantially increasing network size/memory consumption (as it would be if a separate backbone were to be used).

\subsection{Cosine Comparator Head}
\label{sec:cosine_method}
Cosine similarity/distance is defined as
\begin{equation}
  c_i = \cos(\Vec{w_i}, \Vec{x}) = \frac{\Vec{w_i}}{\Vert \Vec{w_i} \Vert} \cdot \frac{\Vec{x}}{\Vert \Vec{x} \Vert}
  \label{eq:cosine}
\end{equation}
where $i = 1, 2, ..., m$ refers to the $i$-th class in an $m$-way classification task. $\Vec{w_i}$ is dynamically computed from exemplar instances via the exemplar pathway in Cos R-CNN's backbone, whereas $\Vec{x}$ is the embedded image feature. In other words, classification is performed in terms of comparison of query to exemplar, and the model can easily adapt to novel classes just by comparing to a novel class exemplar. The L2 normalisation term ensures that the classification weights for both base and novel classes have similar magnitudes as they are bounded to $[-1, 1]$, which allows few-shot classification by exemplars without requiring the model to be fine-tuned. Note that the exemplar embedding module/embedder can alternatively be interpreted as a few-shot weight generator, transforming exemplar features into classification weight vectors (albeit one that is conditional on the input exemplar instead of being a model parameter).

In practice, we replace the fully-connected classification layer in the box class head with a cosine comparator layer, which computes a scaled cosine similarity on query and exemplars,
\begin{equation}
  c'_i = \gamma \cdot \cos(\Vec{w_i}, \Vec{x}) + \beta
  \label{eq:cosine_parameterised}
\end{equation}
where $\gamma$ and $\beta$ are learnable scalar scale and bias parameters. This form allows the model to learn to control the peakiness of the class logits distribution by expanding the domain to $[-\infty, \infty]$ while still having the advantages of a cosine formulation as reported in other works~\cite{NIPS2018_7352,chen2018a}.

As the open background class cannot be represented by exemplars, we still compute the background logits from RoI features using a learnable function, $\phi _{bg}$, which is implemented as a fully-connected layer. We concatenate background logits with the logits computed from exemplars. The combined logits are then softmaxed and passed to a cross-entropy loss objective, which can simply then be backpropagated to train the model. In all, this leads to a definition of the posterior for class $i$ in the $m$-way, 1-shot task as
\begin{equation}
\label{eq:posterior}
  p(y=i|\Vec{x}) = \dfrac{\exp(c'_i)}{\exp(\phi _{bg}(\Vec{x})) + \sum_{j}^m \exp(c'_j)}
\end{equation}

\subsubsection{Few-shot Inference}
So far, we have formulated Cos R-CNN's detection as a comparison between a single example from each class, \ie the one-shot scenario. One of the ways Cos R-CNN can be extended to the $m$-way, $n$-shot detection task is as follows. We provide $n$ exemplars for each of the $m$ classes. Inference then follows as $n \cdot m$-way, 1-shot detection. However, after applying softmax to the cosine-similarity logits across the $n \cdot m$ exemplars, the detection score for each class is computed as the sum of the scores from the $n$ exemplars belonging to the same class. The posterior for class $i$ in the $m$-way, $n$-shot task is thus given by
\begin{equation}
  \label{eq:fewpost}
  p(y=i|\Vec{x}) = \dfrac{\sum^n_k{\exp(c'_{ik})}}{\exp(\phi _{bg}(\Vec{x})) + \sum^m_j\sum^n_k \exp(c'_{jk})}
\end{equation}

It has been found for comparator-based classification that it is beneficial to train with the same number of shots, $n$, as used in testing \cite{NIPS2017_6996}. However, we later evaluate how this formulation of few-shot detection allows for performance improvements that come from additional examples of novel classes, without requiring Cos R-CNN to be re-trained or even fine-tuned.

\subsection{Supervised Momentum Contrast (Su-MoCo)}

It has recently been shown that instance-based contrastive self-supervised learning benefit from being able to iterate on larger batch sizes~\cite{he2019momentum}. As Cos R-CNN is also trained in a contrastive manner (comparison of query to exemplars), we posit that it may also benefit despite it being trained in a supervised manner. However, it is not straightforward to simply increase the batch size as R-CNN architectures use a substantially larger amount of GPU memory due to RoI sampling heuristics (\eg batch sizes of 1 or 2 are commonly used).

One workaround is to serialise exemplar features from previous iterations in a memory bank after detaching these exemplars from the computational graph. Afterwards, in the current training iteration, randomly sample features from this bank and append them as additional instances to the current batch~\cite{Wu_2018_CVPR}. The serialised exemplars consume very little memory and a large number of them may be used to increase the effective "batch size".

However, a drawback to the memory bank is that randomly sampled exemplars might not be coherent as they might have been generated at very different stages of the network evolution during training. MoCo (Momentum Constrast)~\cite{he2019momentum} solves this by adopting a queue strategy (FIFO) for sampling from the memory bank, and using a duplicate of the main network to generate exemplar features that is updated more slowly than the main network. This ensures that the sampled exemplars remain as coherent as possible. We build on this work and introduce a variant for supervised object detection which we call Su-MoCo.

Like MoCo, Su-MoCo also utilizes a slowly-evolving duplicate model for the exemplar pathway and update its weights according to a momentum rule. Namely, we use the same update rule to compute the duplicate model weights
\begin{equation}
\label{eq:momentum_update}
  \theta_k \leftarrow \alpha \theta_k + (1 - \alpha) \theta_q
\end{equation}
where $\theta_q$ are the weights of latest main network in training, and $\alpha$ is the momentum value.

Unlike MoCo, which uses one unsupervised loss for such as InfoNCE~\cite{oord2018representation}, our supervised loss for Su-MoCo is an average of negative log-likelihoods from different queue samples, namely
\begin{equation}
  \mathcal{L} = -\frac{\sum_\ell^q(\log p_\ell)}{q}
\end{equation}
where $p_\ell$ is the posterior in Eq.~\ref{eq:posterior} computed for a particular queue $\ell$ and $q$ is the queue size. Note that because of the R-CNN architecture, each batch of queue posteriors in one iteration of Su-MoCo are sampled from different RoI batches.

Using Su-MoCo allows us to considerably increase the amount of exemplars being compared in one iteration which is beneficial to Cos R-CNN performance as can be seen in \S\ref{sec:ablations}.

\section{Experiments}
\label{sec:experiments}

We now outline experiments that evaluate the performance of Cos R-CNN, and show ablations on its design. Specifically, we benchmark against 2 distinct tests: 5-way ImageNet~\cite{Karlinsky_2019_CVPR} and 20-way Pascal VOC~\cite{Kang_2019_ICCV}. The 2 tests allow us to benchmark our method in both an episodic (as is prevalent in few-shot classification literature) and non-episodic manner.

\subsection{Few-shot ImageNet}
\label{sec:imagenet_test}

We first test our method on the ImageNet 5-way, $n$-shot detection benchmark proposed by \cite{Karlinsky_2019_CVPR}. The benchmark is based on a 314-class subset (containing mostly animal classes) of the ImageNet CLS-LOC dataset~\cite{Russakovsky2015}, partitioned into 100 base and 214 novel classes. After training on the 100 base class examples, the model is tested on the unseen 214 novel classes by evaluating 500 episodes of 5-way, $n$-shot few-shot tasks. Each episode consists of 50 query images (10 images for each of the 5 randomly sampled novel classes for this episode) and $5n$ exemplar images ($n$ exemplars for each of the 5 novel classes). Detection metrics are then reported by jointly evaluating test images from all 500 episodes.

\subsubsection{Model}
Unless otherwise stated, we base our detection model on Faster R-CNN with a ResNet50-FPN backbone~\cite{Lin_2017_CVPR}. We keep the original RPN for region proposal, and instead only modify the classification box head with the cosine classifier outlined in \S\ref{sec:cosine_method} to perform few-shot detection. Consequently, we also use a class-agnostic bounding box regressor in the box head to be able to adapt to any number of classes. For region-of-interest (RoI) pooling and warping, we utilise the more recent RoIAlign instead of RoIPool~\cite{He_2017_ICCV}.

\subsubsection{Training}
We initialise the model parameters by sampling from a Gaussian distribution with $\mu=0$ and $\sigma=0.01$ except for the ResNet50 backbone which was pre-trained on COCO~\cite{10.1007/978-3-319-10602-1_48}, and the cosine comparator where we set $\gamma=1$ and $\beta=0$ (see Eq.~\ref{eq:cosine_parameterised}). We also freeze all the batch normalization layers of the pretrained model and backbone until the second ResNet stage.

We train our model using SGD with a momentum of 0.9 and weight decay of $10^{-4}$ for 240,000 iterations, starting by warming up the learning rate linearly to 0.0375 at iteration 500. The learning rate is then decayed by a factor of 10 at iterations 160,000 and 213,333.

We utilise image-centric episodic training on 3 GPUs with a batch size of 1 image per GPU, where each iteration in each GPU consists of an episode comprising 1 query image and 5 exemplar instances. We resize query images and exemplars so that their shorter side is at least 600 pixels, and obtain exemplar instances after by cropping them from their ground truth bounding boxes. We also flip the images/exemplars horizontally with a probability of 0.5 as a form of data augmentation. Finally, we do not use Su-MoCo here to expand the effective exemplar batch size, as the episodic training format means that exemplars of the previous batch are obselete for the current iteration.

\subsubsection{Inference}
Taking the model at the end of the training schedule, we use the identical 500 episodes (query/exemplar image sets) used by RepMet~\cite{Karlinsky_2019_CVPR} to evaluate the novel subset, and sample our own episodes for the base subset as the exact episodes were not provided. Evaluation on shots $>$ 1 uses per-channel averaging of the exemplar features at the comparison layer.

\subsubsection{Results}
Our results are shown in Table \ref{tab:5way_results}, where we outperform state-of-the-art RepMet~\cite{Karlinsky_2019_CVPR} on both base and novel class detections. Base class detections are evaluated in an episodic manner like novel classes (\ie by comparing to exemplars). Note that we adopt the COCO convention and write our results as AP$_{50}$, which is equivalent to ILSVRC mAP as reported in RepMet.

Table \ref{tab:5way_results} also shows the evaluation of our extension into few-shot detection. We demonstrate that Cos R-CNN is able to improve detection through the combination of multiple examples of each class. It is important to note that Cos R-CNN has only been trained in the one-shot scenario and few-shot detection is performed online (\ie without any fine-tuning or re-training to the new shot number).

\begin{table}
  \begin{center}
    \caption{AP50 scores on 5-way ImageNet LOC test proposed in RepMet~\cite{Karlinsky_2019_CVPR}. We obtain improved results on 1-shot, 5-shot, and 10-shot compared to their results without fine-tuning on novel classes. Both models use the same episodes to evaluate the novel subset performance.}
    \label{tab:5way_results}

    \begin{tabular}{@{}lrrrrrr@{}}
      \toprule
      & \multicolumn{3}{r}{Base AP$_{50}$} & \multicolumn{3}{r@{}}{Novel AP$_{50}$} \\
      \cmidrule(r){2-4} \cmidrule(l){5-7}
      Method/Shot & 1 & 5 & 10 & 1 & 5 & 10 \\
      \midrule
      RepMet & 64.5 & 79.4 & 82.6 & 56.9 & 68.8 & 71.5 \\
      Cosine & \hl{84.5} & \hl{88.4} & \hl{89.4} & \hl{65.1} & \hl{72.3} & \hl{72.9} \\
      \bottomrule
  \end{tabular}
  \end{center}
\end{table}

\subsection{Few-shot VOC}
\label{sec:voc_test}

\begin{table*}
    \begin{center}
      \caption{AP50 scores on 20-way VOC test proposed by FewShot-YOLOv2~\cite{Kang_2019_ICCV}. All models are trained on the base classes and not fine-tuned to the novel class exemplars. The same exemplars obtained from~\cite{Kang_2019_ICCV} were used to evaluate all models. Best novel class results among the same backbone are in bold.}
      \label{tab:20way_results}

      \subfloat{\begin{tabular}{@{}llrrrrrrrrrr@{}}
        \toprule
        \multicolumn{2}{@{}l}{VOC Split 1} & \multicolumn{5}{r}{Base AP$_{50}$/shot} & \multicolumn{5}{r@{}}{Novel AP$_{50}$/shot} \\
        \cmidrule(r){3-7} \cmidrule(l){8-12}
        Method      & Backbone      & 1 & 2 & 3 & 5 & 10 & 1 & 2 & 3 & 5 & 10 \\
        \midrule
        FS-YOLOv2   & DarkNet       &     67.3  &     69.4  &     69.5  &     69.5  &     69.6  &      2.0  &      8.2  &      8.7  &     11.8  &     11.9  \\
        Meta R-CNN  & R-101-FPN     &     68.9  &     68.9  &     69.1  &     69.1  &     69.1  &      2.9  &      3.6  &      3.2  &      3.2  &      3.4  \\
        RepMet      & R-101-DCN     &     68.4  &     67.0  &     70.5  &     70.7  &     72.5  &     29.0  &     32.8  &     30.9  &     35.6  &     40.2  \\
        \midrule
        RepMet      & R-50-FPN      &     65.3  &     68.9  &     67.6  &     69.1  &     70.1  &     14.9  &     11.1  &     13.6  &     19.3  &     21.3  \\
        L2          & R-50 FPN      &     65.8  &     63.9  &     66.0  &     66.4  &     66.4  & \hl{30.2} &     21.9  &     24.2  &     31.4  &     32.0  \\
        Cosine      & R-50-FPN      &     63.4  &     67.2  &     67.6  &     67.7  &     67.5  &     27.9  & \hl{33.0} & \hl{32.1} & \hl{36.2} & \hl{33.6} \\
        \bottomrule
      \end{tabular}}

      \subfloat{\begin{tabular}{@{}llrrrrrrrrrr@{}}
        \toprule
        \multicolumn{2}{@{}l}{VOC Split 2} & \multicolumn{5}{r}{Base AP$_{50}$/shot} & \multicolumn{5}{r@{}}{Novel AP$_{50}$/shot} \\
        \cmidrule(r){3-7} \cmidrule(l){8-12}
        Method      & Backbone      & 1 & 2 & 3 & 5 & 10 & 1 & 2 & 3 & 5 & 10 \\
        \midrule
        FS-YOLOv2   & DarkNet       &     66.2  &     71.2  &     71.4  &     70.1  &     70.0  &     11.9  &      1.0  &      4.6  &      4.1  &      1.1  \\
        Meta R-CNN  & R-101-FPN     &     69.7  &     64.5  &     70.1  &     71.2  &     71.5  &      1.5  &      1.1  &      1.0  &      3.5  &      4.6  \\
        RepMet      & R-101-DCN     &     68.1  &     66.6  &     69.6  &     69.7  &     71.9  &     11.1  &     10.8  &     16.8  &     18.0  &     21.5  \\
        \midrule
        RepMet      & R-50-FPN      &     65.2  &     67.0  &     65.9  &     68.3  &     68.9  &     15.3  & \hl{14.0} &     12.3  &     15.7  &     18.9  \\
        L2          & R-50-FPN      &     64.8  &     64.9  &     66.6  &     66.7  &     66.4  &     14.3  &     10.0  &     14.2  &     13.4  &     18.7  \\
        Cosine      & R-50-FPN      &     62.5  &     66.8  &     67.2  &     67.4  &     67.3  & \hl{19.4} &     12.6  & \hl{14.4} & \hl{19.1} & \hl{21.9} \\
        \bottomrule
      \end{tabular}}

      \subfloat{\begin{tabular}{@{}llrrrrrrrrrr@{}}
        \toprule
        \multicolumn{2}{@{}l}{VOC Split 3} & \multicolumn{5}{r}{Base AP$_{50}$/shot} & \multicolumn{5}{r@{}}{Novel AP$_{50}$/shot} \\
        \cmidrule(r){3-7} \cmidrule(l){8-12}
        Method      & Backbone      & 1 & 2 & 3 & 5 & 10 & 1 & 2 & 3 & 5 & 10 \\
        \midrule
        FS-YOLOv2   & DarkNet       &     68.9 &     70.0  &     70.1  &     70.1  &     70.1  &      6.0  &      7.6  &      8.5  &      9.9  &      9.5  \\
        Meta R-CNN  & R-101-FPN     &     68.9 &     68.9  &     69.1  &     69.1  &     69.2  &      1.7  &      2.9  &      1.4  &      0.8  &      1.1  \\
        RepMet      & R-101-DCN     &     63.3  &     64.0  &     67.2  &     68.5  &     72.2  &     21.1  &     21.7  &     24.9  &     26.5  &     27.9  \\
        \midrule
        RepMet      & R-50-FPN      &     63.4  &     62.8  &     55.1  &     66.1  &     68.4  &     16.0  &     12.4  &      2.0  &     15.8  &     14.4  \\
        L2          & R-50-FPN      &     60.1  &     64.8  &     65.0  &     65.8  &     65.5  & \hl{17.1} &     14.8  &     20.4  &     22.3  &     20.4  \\
        Cosine      & R-50-FPN      &     59.7  &     66.1  &     66.9  &     67.1  &     67.0  &     16.9  & \hl{21.6} & \hl{21.6} & \hl{27.5} & \hl{25.5} \\
        \bottomrule
      \end{tabular}}
    \end{center}
  \end{table*}

Next, we evaluate against the 20-way, $n$-shot detection benchmark by \cite{Kang_2019_ICCV}. The benchmark is based on the PASCAL VOC object detection dataset~\cite{Everingham15} and consists of 3 different splits consisting of 15 base and 5 novel classes each. Unlike few-shot ImageNet (which can form many episodes of different class combinations), this testing is non-episodic and we report the results on each split directly: after training on the 15 base class examples, the model is tested on the unseen 5 novel classes. Again, we report AP$_{50}$ as our evaluation metric.

\subsubsection{Training}
The same training hyperparameters and schedules were used as in \S\ref{sec:imagenet_test}, except that a) we now pre-train the backbone on ImageNet as COCO and PASCAL VOC classes overlap, and b) each iteration consists of 1 query image and 15 exemplar instances. Additionally, we utilize Su-MoCo with a queue size of 100 and momentum of 0.999 to boost the effective number of exemplars seen in each iteration by a factor of 100. As the ablations in \S\ref{sec:ablations} show, using Su-MoCo helps improve the final performance of the model.

\subsubsection{Inference}
For all splits, we use the same exemplars provided by~\cite{Kang_2019_ICCV} to evaluate the base and novel class metrics using the model obtained at the end of the training schedule. Like \S\ref{sec:imagenet_test}, evaluation on shots $>$ 1 uses per-channel averaging of the exemplar features at the comparison layer.

\subsubsection{Baselines}
We also trained FewShot-YOLOv2~\cite{Kang_2019_ICCV}, Meta R-CNN~\cite{Yan_2019_ICCV}, and RepMet~\cite{Karlinsky_2019_CVPR} using code provided by each respective authors and their original schedules, but without fine-tuning to few-shot exemplars as we want to evaluate their suitability under an online setting. To better control for differences in model architecture, we also trained a ResNet-50 variant of RepMet without deformable convolutions~\cite{Dai_2017_ICCV}, OHEM~\cite{Shrivastava_2016_CVPR}, or Soft-NMS~\cite{Bodla_2017_ICCV} using our training schedule to isolate the difference between both methods to just their metric learning formulations. Finally, we also show results where we substituted the cosine similarity function with an Euclidean distance (L2) function ala Prototypical Networks~\cite{NIPS2017_6996}.

\subsubsection{Results}
We report our results and baselines on base and novel classes in Table~\ref{tab:20way_results}. While FS-YOLOv2 and Meta R-CNN were meta-learned, they were not designed to do online detection, which is reflected in their poor performance on the novel unseen classes. In contrast, the original RepMet implementation on R-101-DCN is comparable or better than any of the R-50-FPN variants shown in the table. However, this edge may be attributed to a deeper backbone and/or bells-and-whistles, as we see that on average, the simpler cosine formulation outperforms the vanilla R-50-FPN variant of RepMet on the novel classes by a large margin.

We also notice that the base class performance of cosine is very slightly worse compared to other methods. However, the primary objective of an online few-shot method is to perform well on the novel classes, and we interpret this as a reasonable trade-off to make for pushing novel class performance.

We also show results of using the Euclidean distance as the comparison metric, and we see that it is slightly worse on average than using cosine, highlighting the importance of using cosine.

\subsection{Ablations}
\label{sec:ablations}

Unless otherwise stated, we detail ablations on Cos R-CNN by comparing results on 5-way ImageNet detection. For these ablations, we use a slightly different episodic testing format: we evaluate all images by forming $N$ episodes of 1 query image each, where $N$ is the number of images in the dataset. This allows for quicker evaluation without losing comprehensiveness. Some ablations are based on few-shot VOC, and for these ablations we report novel class results on the first split.

\subsubsection{Cosine Form}
Here we examine the necessity of applying a parameterised scalar affine transformation (in the form of scale $\gamma$ and bias $\beta$) on the cosine similarity output. Doing this frees the constraint that logits have to lie between $[-1, 1]$ into $[-\infty,\infty]$, while still keeping the advantages of a cosine-based formulation. Results on Table \ref{tab:ablations_comparator} confirm this intuition. We observe that the scale factor $\gamma$ is the most influential component and should almost always be used in conjunction with cosine similarity. In the presence of $\gamma$, the bias factor $\beta$ seems to have less of an effect, and in the case above, slightly degrades detection performance.

\subsubsection{Loss Objective}
In Table \ref{tab:ablations_softmax}, we further compare learning the model through a cross-entropy (softmax) vs. a binary cross-entropy (sigmoid) objective, which is a common alternative formulation used in other detectors such as RetinaNet~\cite{8237586}. For binary cross-entropy, we do not directly compute background logits, but instead rely on the class detection scores being low for background RoIs. While in standard non-few-shot detection pipelines either formulation works, we find that for cosine-similarity based few-shot detection, a softmax loss gives superior results. Observing the detection results, we find there is a tendency to have predictions of all exemplar classes for the same box, which brings down the detection performance by reducing the classification rate and suppression of background classes.

\begin{table}
    \begin{center}
      \caption{Ablations on the cosine box head reported as AP$_{50}$.}
      \label{tab:ablations_cosine}
      \subfloat[Cosine form]{
        \label{tab:ablations_comparator}
        \begin{tabular}{@{}lrr@{}}
          \toprule
          Form & Base & Novel \\
          \midrule
          No scale or bias & 49.2 & 35.3 \\
          Scale & \hl{85.9} & \hl{65.2} \\
          Scale and bias & 85.5 & 64.5 \\
          \bottomrule
        \end{tabular}
      }
      \subfloat[Loss objective]{
        \label{tab:ablations_softmax}
        \begin{tabular}{@{}lrr@{}}
          \toprule
          Loss & Base & Novel \\
          \midrule
          Softmax & \hl{85.5} & \hl{64.5} \\
          Sigmoid & 45.4 & 48.3 \\
          \midrule
        \end{tabular}
      }
    \end{center}
  \end{table}

\subsubsection{Cos RPN}

\begin{table}
    \begin{center}
      \caption{Cos RPN embedder ablations reported as AP$_{50}$.}
      \label{tab:ablations_rpn}
      \subfloat{
        \label{tab:ablations_rpn_voc}
        \begin{tabular}{@{}lrrrrr@{}}
          \toprule
          Method/Shot & 1 & 2 & 3 & 5 & 10 \\
          \midrule
          Standard & 27.9 & \hl{33.0} & \hl{32.1} & \hl{36.2} & 33.6 \\
          Cos (Linear) & \hl{30.6} & 29.1 & 29.9 & 33.6 & \hl{34.2} \\
          \bottomrule
        \end{tabular}
      }
    \end{center}
  \end{table}

One possible design decision is to also apply our cosine similarity formulation to the RPN, modifying the standard RPN objectness classifier by taking the maximum detection score at any point across all exemplars as the objectness score.

In Table \ref{tab:ablations_rpn_voc}, we show the results of a standard RPN against a Cos RPN. The Cos RPN is a slight modification to the exemplar embedding function within the RPN. We use a single fully-connected layer as the embedder function.

From the overall trends, Cos RPN appears to be less optimal compared to standard RPN for few-shot detection. A possible explanation is that object appearances are too varied to reliably cluster around a single centre, and so a cosine similarity formulation would be less able to capture all possible modes of variation. Alternatively, it is possible that the diverse variety of object appearances in the base training data is sufficient to enable the standard RPN propose good bounding boxes of unseen objects. As such, we adopt standard RPN in our main experiments.

However, this does not mean that the cosine similarity in the box head is superfluous. It is also possible that while standard RPN in a standard R-CNN system predicts objectness, standard RPN in Cos RPN could also be benefitting from gradients backpropagated by the cosine box head which allows it to generate good proposals of novel class objects. Additionally, without the cosine formulation, online detections of novel classes will not be possible.

\subsubsection{Su-MoCo}

\begin{table*}
    \begin{center}
      \caption{Ablations on Su-MoCo AP$_{50}$ perfomance on Split 1 of few-shot VOC.}
      \label{tab:ablations_moco}
      \subfloat[Across hyperparameters]{
        \label{tab:ablations_moco_params}
        \begin{tabular}{@{}llrr@{}}
          \toprule
          Momentum & Queue size & Base & Novel \\
          \midrule
          0.0 & 0 & \hl{65.4} & 26.0 \\
          0.9 & 10 & 64.9 & 29.8 \\
          0.99 & 10 & 64.4 & 27.7 \\
          0.99 & 100 & 62.9 & 28.9 \\
          0.999 & 100 & 62.4 & \hl{30.6} \\
          \bottomrule
        \end{tabular}
      }
      \subfloat[Across $n$-shots]{
        \label{tab:ablations_moco_voc}
        \begin{tabular}{@{}llrrrrr@{}}
          \toprule
          RPN & Method/Shot & 1 & 2 & 3 & 5 & 10 \\
          \midrule
          \multirow{2}{*}{Standard} & Standard & 27.7 & 30.0 & 26.7 & 32.2 & 31.3 \\
          & Su-MoCo & \hl{27.9} & \hl{33.0} & \hl{32.1} & \hl{36.2} & \hl{33.6} \\
          \midrule
          \multirow{2}{*}{Cos (Linear)} & Standard & 26.0 & 27.5 & 27.4 & 31.9 & 31.6 \\
          & Su-MoCo & \hl{30.6} & \hl{29.1} & \hl{29.9} & \hl{33.6} & \hl{34.2} \\
          \bottomrule
        \end{tabular}
      }
    \end{center}
  \end{table*}

We first ablate the momentum and queue size hyperparameters for Su-MoCo as shown in Table~\ref{tab:ablations_moco_params}. Momentum controls how quickly the exemplar pathway weights evolve during training, and the queue size determines the effective size of exemplars evaluated in a training iteration. We find performance on novel classes improves as we increase queue size, confirming the intuition that the model learns better feature representations if it can consider more exemplars during a training iteration. However, selecting an appropriate momentum value for the selected queue size is also important. As queue sizes increase, momentum also needs to increase as the exemplar pathway weights need to evolve more slowly in order to ensure old exemplars in the queue still remain informative~\cite{he2019momentum}. We find that a good rule is to set the momentum value to  $1 - q^{-1}$ where $q$ is the queue size. As we find $q = 100$ to work best for novel classes, we use this value and thus a momentum of 0.999 for our experiments in \S\ref{sec:voc_test}.

These findings on queue size corroborate~\cite{he2019momentum}, but we find that for supervised learning (our setting), the queue sizes do not have to be extremely large in order to observe concrete improvements in performance.

Finally, in Table~\ref{tab:ablations_moco_voc}, we demonstrate the effectiveness of adopting Su-MoCo, where we obtain improvement of 2-4\% across the board on novel classes compared with not using Su-MoCo.

\section{Conclusion}
\label{sec:conclusion}

In this paper, we have proposed the Cos R-CNN online few-shot object detector, and validated its design through ablation studies. The performance of Cos R-CNN was evaluated on a recently introduced ImageNet 5-way few-shot detection benchmark, beating the state-of-the-art in 1/5/10-shot setups by 8/4/1\% whilst also being a simpler formulation. This is also reflected in the few-shot PASCAL VOC benchmark, beating contemporary methods on the novel classes by up to 20\%.

\bibliographystyle{IEEEtran}
\bibliography{IEEEabrv,library}

\end{document}